\documentclass[runningheads,a4paper]{llncs}

\usepackage{amssymb}
\usepackage{amsmath}
\usepackage{graphicx}
\usepackage{url}
\usepackage{natbib}

\usepackage[utf8]{inputenc}

\begin{document}

\mainmatter

\title{Generalised Random Forest Space Overview}
\titlerunning{Generalised Random Forest Space Overview}

\author{Miron B. Kursa}

\authorrunning{M.B. Kursa}

\institute{Interdisciplinary Centre for Mathematical and Computational Modelling (ICM), University of Warsaw, \\ Pawi\'{n}skiego 5A,
02-106 Warsaw, Poland\  \\
\email{M.Kursa@icm.edu.pl}}

\toctitle{Generalised Random Forest Space Overview}
\tocauthor{Miron B. Kursa}
\maketitle


\begin{abstract}
Assuming a view of the Random Forest as a special case of a nested ensemble of interchangeable modules, we construct a generalisation space allowing one to easily develop novel methods based on this algorithm.
We discuss the role and required properties of modules at each level, especially in context of some already proposed RF generalisations.
\end{abstract}

\section{Introduction}
Random Forest (RF) is a popular, powerful ensemble machine learning method proposed by \citet{breiman_random_2001}.
Although the canonical version of this algorithm is known to be very versatile and perform well in numerous applications, many variants of this method have been proposed, for many different purposes: to extend the RF capabilities, to generalise over specific, non-standard data, to increase accuracy in a certain conditions,  to improve the attribute importance measure produced by this method, to speed-up training or prediction, just to name a few.

This work aims to provide a conceptual framework of generalised Random Forest (GRF) methods, useful both in classification of existing RF variants and defining research opportunities in this field.

\section{Generalised Random Forest}
\begin{figure}
 \centering
 \includegraphics[width=\textwidth]{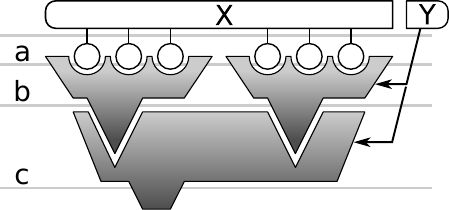}
 \caption{\label{fig:genGRF} Structure of a Generalised Random Forest. a)~pivots, b)~sharpening ensemble, c)~conditioning ensemble. X and Y denote, respectively, predictor and decision part of the information system.}
\end{figure}

In the presented model of the generalised Random Forest, we assume a following three-layer nested ensemble structure, as shown on Figure~\ref{fig:genGRF}.

The data is ingested into the model via numerous \textit{pivot models}, which have to fit the form of the data and are expected to be very simple and easy to train, although are not necessary have to be neither accurate nor robust; their intended role is to become an interface between the input and internal logic of the GRF model.
In other words, we assume that they handle the feature construction step of the modelling process.

Pivot models are grouped are converted into a meaningful ensemble models with the \textit{sharpening ensembles}; this modules of this class orchestrate the generation of pivots by using the information from the decision attribute.
The sharpening ensemble is expected to produce accurate models regardless of their robustness; however, they must not require external optimisation of any kind and should be computationally efficient.

The outermost layer of the GRF is the \textit{conditioning ensemble} which builds and groups sharpening ensembles.
Its role is to finally build a robust model by joining a lot of accurate models for which one does not know which are overfitted and which are not.
It is trivial to show that this can be achieved even with a very small fraction of meaningful models, provided that sharpening ensembles will not be correlated; to this end the conditioning ensemble usually involves some permutation strategy.

Moreover, conditioning ensemble can be expanded to provide a generalisation of some of the additional features of Random Forest, like internal error approximation, object-object dissimilarity measure or feature importance.
It is also a good place to implement parallel computing capabilities.

\section{Pivot models}
Pivot models are the only part of the GRF structure that has a direct contact with predictors of an information system, thus it is a place for modifying the interface with the data structure.
The form of an output of a pivot classifier and the algorithm which is used to train it is naturally enforced by a form of the sharpening ensemble; most often pivot forms an embranchment in some kind of a decision tree, thus shall return a direction in which certain object should descent within the tree.
As the trees are most often binary, this gives us two options, which we will later in the paper call $R$ (for right) and $L$ (for left).

In a standard Random Forest \citep{breiman_random_2001}, it is assumed that an information system is composed of either categorical or continuous\footnote{One should note that ordered categorical data, like for instance $\text{cold}<\text{warm}<\text{hot}$, can be treated as continuous without any loss of generality.} predictors.
This way, we have only two possible pivot classifiers, respectively for categorical and continuous feature, in a following forms:
\begin{equation}
 f(x\in \mathcal{C}_{x};\Xi\in2^{\mathcal{C}_{x}})\rightarrow \{R,L\}:=\left\{
  \begin{array}{l}
   R:x\in\Xi,\\
   L:x\notin\Xi,
  \end{array}
 \right.,
\end{equation}
where $\mathcal{C}_{x}$ is a set of categories of $x$, and
\begin{equation}
 f(x\in \mathbb{R};\Xi\in\mathbb{R})\rightarrow \{R,L\}:=\left\{
  \begin{array}{l}
   R:x\geq \Xi,\\
   L:x<\Xi,
  \end{array}
 \right..
\end{equation}
In either case, $x$ is a feature on which the pivot is executed, and $\Xi$ is the \textit{threshold} value.
Both are selected when pivot is generated within the sharpening ensemble; most often they are optimised to maximise the decision homogeneity within the subsets of objects sent left and right, usually measured by information gain or Gini index.

The other popular idea, started by \citet{geurts_extremely_2006} with their Extra-Trees method, is to generate pivots at random, obviously with constraint that a resulting criterion should not sent all objects in one direction.
Such a method removes a problem of finding an appropriate homogeneity measure and performing the optimisation, which in return increases the computational efficiency and, in a number of problems, the robustness of a final model as it leads to a greater divergence among sharpening ensemble models.
Naturally, this way one also removes an impact of a possible problems with the homogeneity measure, which may surface for instance in case of unbalanced sets.
Still, there are problems in the probability of finding even a slightly meaningful pivot is very small, and they usually case random pivot-based algorithm to perform poorly.

And in-between solution is to use some kind of heuristic instead of a full optimisation; a simple example of this technique is to generate some number of random pivots and select the best one.
Other approaches include generic soft optimisation methods like genetic algorithms, randomly reducing the search space (often the set of used predictors, ad done by the standard Random Forest) or disturbing the homogeneity measure.
\citet{rodriguez_rotation_2006} proposed to employ PCA in pivot generation.

The other way of generalising the pivot models is to modify their structure to accommodate information systems going beyond simple sets of categorical and continuous predictors.
\citet{bosch_image_2007} perform image classification with GRF with pivots calculating two vector image descriptors, summing their valus with random weights and comparing with a random threshold.
\citet{fan_kernel-induced_2009} proposed to employ some kernel function used for SVM and build pivot by partial application of this kernel to a randomly selected object and compare the result with a threshold optimised in terms of information gain.
This approach was later used for sound and temporal gene expression patterns \citep{cao_signal_2010,fan_functional_2010}.

\section{Sharpening ensemble}
\begin{figure}
 \centering
 \includegraphics[width=\textwidth]{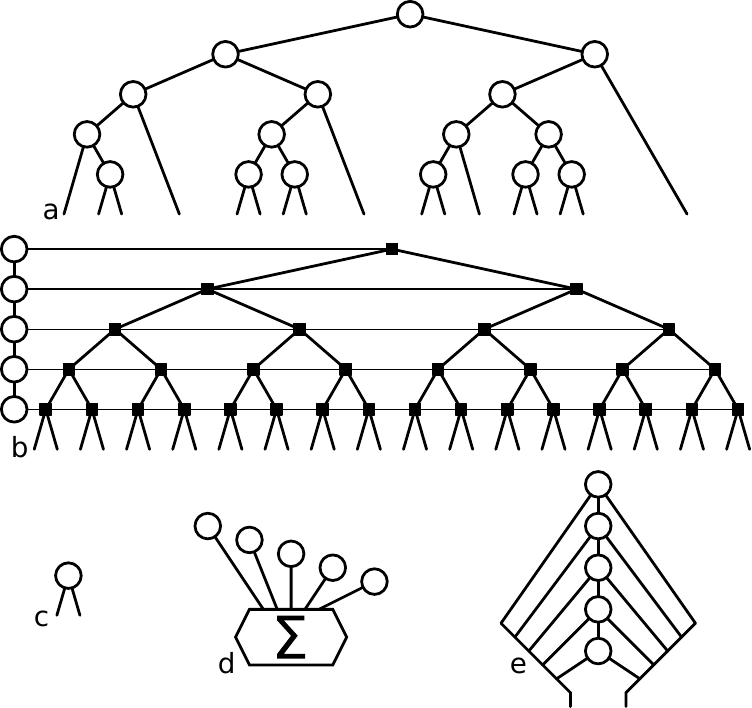}
 \caption{\label{fig:varSE} Examples of sharpening ensemble structures. Embranchment pivot based: a)~decision tree, b)~fern, c)~null ensemble; generic: d)~boosting, e)~decision trunk.}
\end{figure}

The selection of a sharpening module is more complex, since it gives a lot of space for different solutions and approaches.
Some of a popular options, illustrated in Figure~\ref{fig:varSE}, are:
\begin{itemize}
 \item \textbf{decision tree} is a sharpening module used by the Random Forest method; it uses simple embranchment pivots stacked in an iterative manner, i.e. after applying a pivot a sub-tree is build separately for each of the branches, until the sets of objects in leaves will become homogeneous or some pre-set maximum tree depth will be achieved.
 Training of pivots within a decision tree may be performed both randomly or via optimisation; the leaves of a tree will likely end up homogeneous either way.
 To this end, the output of a tree sharpening module is a direct prediction of class.
 \item \textbf{decision fern} \citep{ozuysal_fast_2008,kursa_rferns:_2014} is basically a form of a full decision tree in which each pivot module at a given depth is identical.
 To this end, evaluation of a given pivot is not dependent on the others and can happen in any order, which makes a fern more computationally effective than a decision tree.

 On the other hand, this makes optimisation of a fern is highly non-obvious, thus individual pivots are usually generated randomly.
 This way the leaves of a fern are often non-homogeneous, and thus the prediction of a fern is often encoded as a vector of class probabilities, which naturally requires appropriate adaptation of the voting scheme present in the conditioning ensemble.

 \item \textbf{decision trunk}, proposed by \citet{ulfenborg_classification_2013}, is composed of a flat series of pivot models similar to a fern, though  requires the decision to be binary (say $A$ or $B$), and employs pivot modules (\textit{segments}) which classify into three groups, $A$, $B$ meaning that it is certain that an object belongs to a respective class, and $?$ meaning that the decision is relayed to a next pivot classifier within the trunk.
 Obviously, different than in case of decision ferns, the order of trunk segments is significant because the consideration at level $i$ comes in a context that the object was claimed undecided by segments $1,\ldots,i-1$\footnote{This way trunks can be perceived as a discrete version of boosting.}.
 Trunks practically cannot be implemented without optimisation of pivot models and thus they provide sharp predictions similar to decision trees.

 One should note that ternary pivot models can be realised by combining a pair of regular entrancement pivots, only modified to optimise homogeneity in one branch, respectively for class $A$ and $B$; $?$ is then given to objects directed to second branch in both pivots and for those for which prediction of both pivots are in conflict.

 \item Although \textbf{boosting} \citep{schapire_strength_1990} is almost always considered as a stand-alone ensemble, wrapping it in another layer may lead to a better resilience to noise and mislabelled objects.
 Moreover, boosting has a good support for regression problems, making it a promising alternative to regression trees in context of some of their known problems in this set-up.

 \item there is also a degenerate option which we will call here \textbf{null ensemble}, i.e. just using a single pivot classifier.
 This solution can be effective in case when pivot classifiers are very good on their own and applying a more complex sharpening will only result in an increased computational load.
 A litmus test for such a situation is when a more complex sharpening ensemble of a dynamic complexity, such as a decision tree, is creating shallow models composed of a small number of pivots.
\end{itemize}

Moreover, sometimes the procedure of building the sharpening ensemble is modified to support the outer conditioning ensemble in de-correlation of individual sharpening ensembles.
For instance, in the canonical Random Forest each pivot is build on a randomly generated subset of attributes; this approach is very generic and can be easily ported over other sharpeners, yet obviously a number of alternative methods can be applied as well.

\section{Conditioning ensemble}
As mentioned earlier, the role of the conditioning ensemble is to justify a robust prediction from a set of sharpening ensembles of an unknown reliability.
To this end, this module has to either somehow assess the member models or ensure independence of members so that the noise generated by the overfitted ones will average-out during voting.
The first approach is yet rarely used, mostly because having a reliable enough method of assessing robustness would in practice mean that employing the whole GRF structure is redundant.

The second approach is mostly, as in the canonical Random Forest algorithm, realised through \textit{bootstrap aggregation} (\textit{bagging}), also proposed by \citet{breiman_bagging_1996}, or some variation of this method.
Precisely, the procedure generates a number of object sub-samples for each of the sharpening ensemble; this way they are presented with differently stressed incarnations of the training data and thus exploring a wider range of aspects and becoming less correlated.
Similar approach can be applied to features or implemented as weighting instead of strict selection.
The other substantial option is to simply use a very variable sharpener that will generate diverse models on its own.

\section{Conclusions}
By re-imagining Random Forest as a three-level nested ensemble, we propose a generic, modular framework for extending and modifying this method.

\subsubsection*{Acknowledgments.} This work has been financed by the National Science Centre, grant 2011/01/N/ST6/07035.

\bibliographystyle{plainnat}

\end{document}